%% file: root.tex
\documentclass[letterpaper, 10 pt, conference]{ieeeconf}  

\IEEEoverridecommandlockouts                              

\usepackage{graphicx} 
\usepackage{epsfig} 
\usepackage{mathptmx} 
\usepackage{times} 
\usepackage{amsmath} 
\usepackage{amssymb}  
\usepackage{hyperref}
\usepackage{booktabs}
\usepackage{threeparttable}

\usepackage{pifont}  
\usepackage{capt-of}

\title{\LARGE \bf
V-Dreamer: Automating Robotic Simulation and Trajectory Synthesis via Video Generation Priors
}

\author{Songjia He$^{1*}$, Zixuan Chen$^{1*}$, Hongyu Ding$^{1}$, Dian Shao$^{2}$, Jieqi Shi$^{1}$, Chenxu Li$^{1}$, Jing Huo$^{1}$, Yang Gao$^{1}$
\thanks{$^{*}$These authors contributed equally to this work.}%
\thanks{$^{1}$Nanjing University.}%
\thanks{$^{2}$Northwestern Polytechnical University.}%
}
\makeatletter
\let\@oldmaketitle\@maketitle
\renewcommand{\@maketitle}{%
    \@oldmaketitle
    \centering
    \setcounter{figure}{0}
    \includegraphics[width=\linewidth]{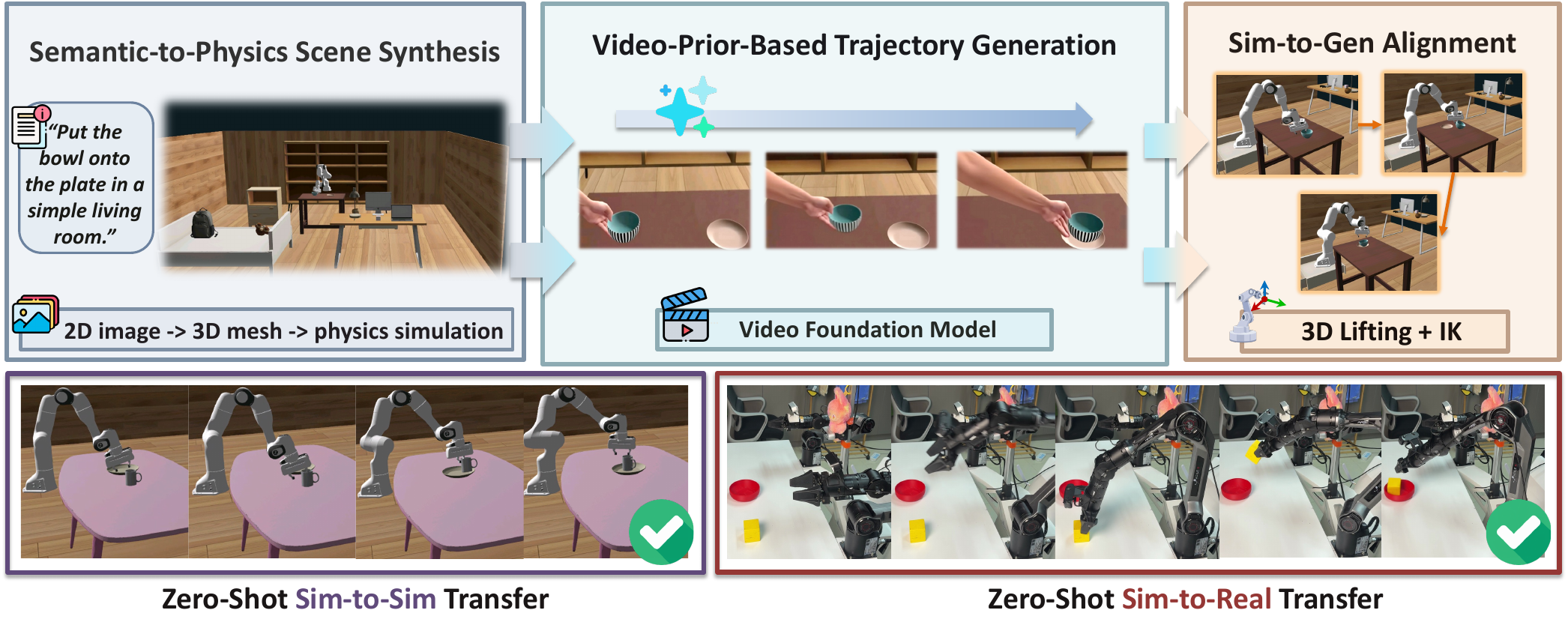}
    \captionof{figure}{We propose \textbf{V-Dreamer}, a fully automated full-cycle pipeline for instruction-driven, open-vocabulary robotic manipulation data synthesis. Given a natural language instruction (optionally with a real-scene photo), V-Dreamer generates a simulation-ready scene and an executable expert trajectory, which are then used to train a policy that can zero-shot generalize to unseen objects in simulation and transfer zero-shot to real robotic hardware.}
    \label{fig:teaser}
}
\makeatother

\begin{document}

\maketitle
\thispagestyle{empty}
\pagestyle{empty}


\input{section/1_abs}

\input{section/2_intro}
\begin{figure*}[t!]
    \centering
    \includegraphics[width=\linewidth]{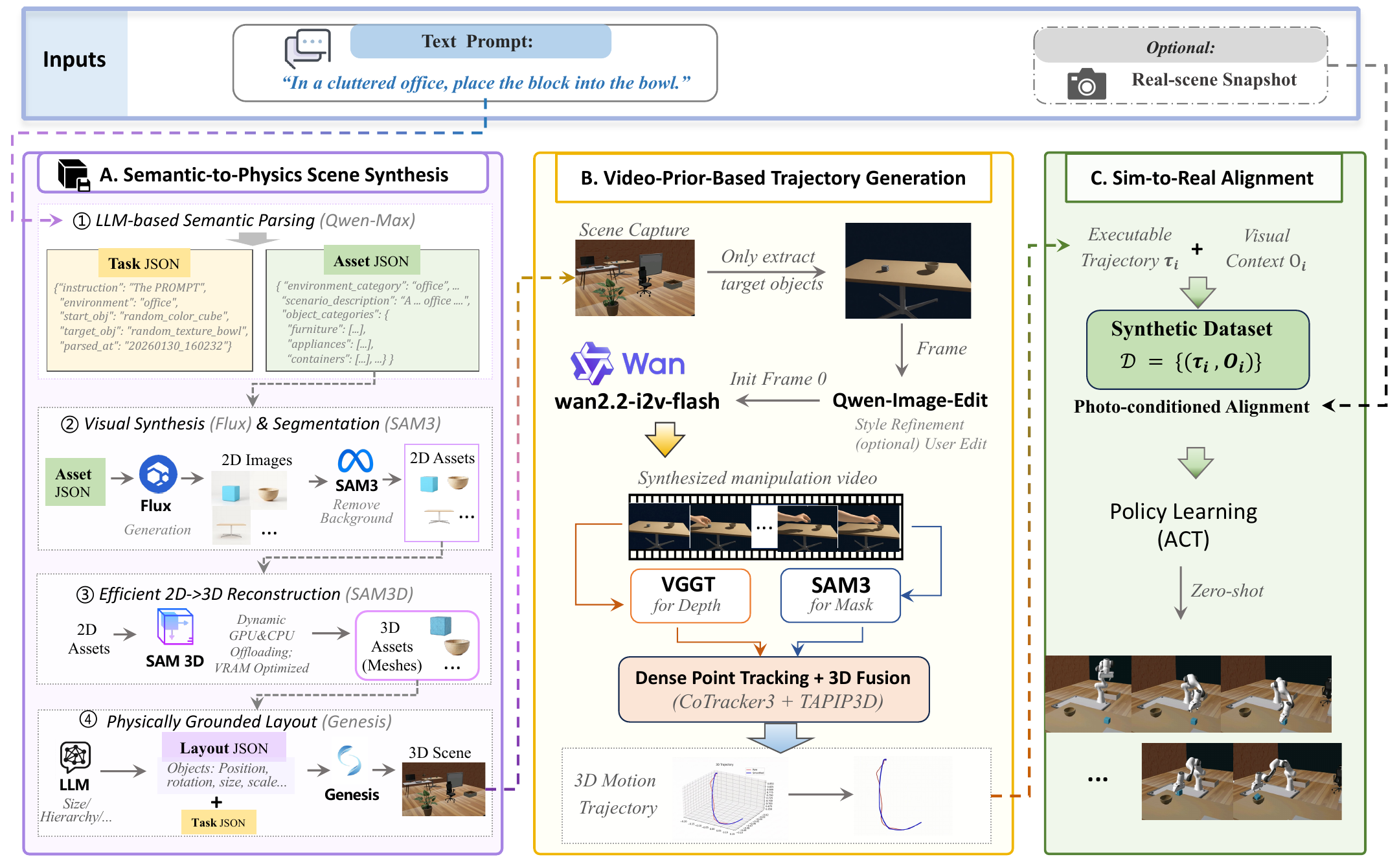}
    \caption{\textbf{Overview of the V-Dreamer Pipeline.}
 The framework consists of three integrated stages: (1) Semantic-to-Physics Scene Synthesis. An LLM-based semantic planner decomposes prompts into asset manifests, which are then transformed into 3D meshes via generative visual synthesis and memory-efficient reconstruction before being assembled into a physics-validated layout. (2) Video-Prior-Based Trajectory Generation. Using the stabilized scene as a prior, we employ video foundation models with targeted negative prompting to dream up physically plausible manipulation motions. (3) Sim-to-Gen Alignment. Actionable 3D trajectories are extracted from the 2D pixel domain through mask-restricted tracking, metric depth estimation, and 3D motion lifting, ultimately mapping visual dreams into executable robot end-effector poses.}
    \label{fig:pipeline}
    \vspace{-1em}
\end{figure*}
\input{section/3_related}

\input{section/4_method}

\input{section/5_exp}
\input{section/6_conclusion}








{\small
\bibliographystyle{IEEEtran}
\bibliography{reference}
}

\end{document}

%% file: section/1_abs.tex
\begin{abstract}

Training generalist robots demands large-scale, diverse manipulation data, yet real-world collection is prohibitively expensive, and existing simulators are often constrained by fixed asset libraries and manual heuristics. To bridge this gap, we present V-Dreamer, a fully automated framework that generates open-vocabulary, simulation-ready manipulation environments and executable expert trajectories directly from natural language instructions. V-Dreamer employs a novel generative pipeline that constructs physically grounded 3D scenes using large language models and 3D generative models, validated by geometric constraints to ensure stable, collision-free layouts. Crucially, for behavior synthesis, we leverage video generation models as rich motion priors. These visual predictions are then mapped into executable  robot trajectories via a robust Sim-to-Gen visual-kinematic alignment module utilizing CoTracker3 and VGGT. 
This pipeline supports high visual diversity and physical fidelity without manual intervention. To evaluate the generated data, we train imitation learning policies on synthesized trajectories encompassing diverse object and environment variations. Extensive evaluations on tabletop manipulation tasks using the Piper robotic arm demonstrate that our policies robustly generalize to unseen objects in simulation and achieve effective sim-to-real transfer, successfully manipulating novel real-world objects.  Project page: \href{https://jia-handsome.github.io/v-Dreamer/}{https://jia-handsome.github.io/v-Dreamer/}.
\end{abstract}

%% file: section/2_intro.tex
\section{Introduction}

General-purpose robotic manipulation in unstructured environments requires policies capable of generalizing across diverse objects, scenes, and semantic goals. While recent advances in imitation learning have demonstrated impressive capabilities in acquiring complex behaviors from expert demonstrations~\cite{chi2023diffusion, zhao2023learning}, their performance is fundamentally bottlenecked by the scale and diversity of the training data. Collecting large-scale real-world datasets via teleoperation is prohibitively expensive, time-consuming, and difficult to scale to long-tail scenarios~\cite{brohan2022rt, padalkar2023open}. Although simulation offers a scalable alternative, existing simulators typically rely on fixed asset libraries and manually designed tasks, restricting the semantic and visual diversity necessary for robust sim-to-real transfer.

The emergence of generative foundation models offers a promising avenue to address this data scarcity. Large Language Models (LLMs) and diffusion models encapsulate extensive world knowledge regarding object semantics, spatial relationships, and visual affordances~\cite{yang2023foundation, wang2023voyager}. 
While recent works have explored using foundation models to generate static 3D scenes~\cite{deitke2023procthor} or high-level task plans~\cite{ahn2022can}, they rarely yield a fully executable robotic task. Video generation can produce visually plausible motion, yet often violates geometric consistency or kinematic constraints due to limited physical grounding~\cite{wu2023tune}. The critical bottleneck is to automatically synthesize dynamic, physically feasible manipulation trajectories without human supervision.

To bridge this gap, we introduce \textbf{V-Dreamer}~\ref{fig:teaser}, a fully automated framework that synthesizes open-vocabulary, simulation-ready manipulation environments and executable expert trajectories directly from natural language instructions. V-Dreamer addresses the dual challenges of scene diversity and behavioral realism through a unified generative pipeline. 
First, our \textit{Semantic-to-Physics Scene Generation} module utilizes LLMs and 3D generative models to construct physically grounded 3D scenes. By integrating geometric reasoning with physics-based validation, the system ensures stable, collision-free object layouts that accurately reflect the user's semantic intent.
Crucially, for behavior synthesis, we introduce a novel \textit{Video-Prior-Based Trajectory Generation} approach. Building on the previously generated simulation scene, we leverage state-of-the-art video generation models as rich motion priors to predict the manipulation process, instead of relying on hand-crafted heuristics or traditional motion planners that often lack semantic nuance.
To ground these generated visual motions into actionable data, we develop a robust \textit{Sim-to-Gen} visual-kinematic alignment module utilizing CoTracker3~\cite{karaev2023cotracker} and VGGT~\cite{wang2025vggt}. This module extracts physically valid end-effector trajectories, allowing us to generate diverse, policy-compliant demonstrations at scale across a wide range of object categories and interaction dynamics. Overall, V-Dreamer offers a fully automated pipeline that turns natural language instructions into simulation-ready, physically grounded scenes and executable expert trajectories. By combining physics-validated scene generation with video-model motion priors and robust visual-kinematic alignment, it achieves both open-vocabulary scene diversity and behavior realism, enabling scalable, policy-compliant demonstrations across diverse objects and interaction dynamics without human supervision.

In practice, V-Dreamer serves as a high-throughput data engine, generating approximately \textbf{600} LeRobot-formatted trajectories \textbf{per hour} on an \textbf{8$\times$RTX 4090} workstation. This rapid generation bypasses the bottleneck of slow manual teleoperation, enabling highly efficient policy training. Furthermore, extensive evaluations on tabletop manipulation tasks using a Piper robotic arm demonstrate that policies trained on this synthetic data robustly generalize to unseen objects in simulation. To further bridge the reality gap, V-Dreamer also support a photo-conditioned setting that instantiates a scene-matched simulation directly from real-world images and a language instruction, making the real2sim2real pipeline highly practical. By training on just a single synthesized demonstration from this pipeline, we achieve effective \textbf{zero-shot sim-to-real transfer}. This highlights the immense potential of generative video priors for scalable and generalizable robot learning.

In summary, our main contributions are:
\begin{itemize}
    \item We propose \textbf{V-Dreamer}, which, to our knowledge, is the first fully automated full-cycle pipeline for instruction-driven open-vocabulary data synthesis. By distilling natural language into simulation-ready environments and expert trajectories, V-Dreamer enables scalable policy learning without human-in-the-loop effort.
    \item We introduce a novel \textbf{video-prior-based behavior synthesis} method. By leveraging a robust visual-kinematic alignment module, we successfully ground generative visual motion into physically feasible end-effector trajectories, eliminating the reliance on hand-crafted heuristics.
    \item We empirically validate V-Dreamer as a high-throughput data engine capable of supporting robust zero-shot generalization and highly data-efficient one-shot imitation learning. Notably, via a zero-shot real2sim2real workflow, a policy trained on a single synthesized demonstration achieves successful sim-to-real transfer on physical hardware while remaining robust to local spatial perturbations.
\end{itemize}

%% file: section/3_related.tex
\section{Related Work}

\subsection{Generative Simulation for Robotics}
Simulation is a cornerstone for scaling robot learning, enabling safe and rapid data collection~\cite{todorov2012mujoco, makoviychuk2021isaac}. Early approaches relied on procedural generation with fixed asset libraries to create diverse environments~\cite{deitke2023procthor, szot2021habitat}, but these methods are inherently limited by the diversity of hand-crafted asset databases and often lack photorealistic textures. Recent works have begun to leverage generative models to automate simulation design at a higher level. GenSim~\cite{wang2024gensim} and RoboGen~\cite{wang2023robogen} utilize LLMs to generate task codes and scene configurations, Holodeck~\cite{yang2024holodeck} compositionally arranges assets into 3D environments from language descriptions, and MimicGen~\cite{mandlekar2023mimicgen} automatically scales demonstrations to new object placements. However, all these methods are fundamentally constrained by fixed asset repositories, limiting visual variety and semantic open-endedness. V-Dreamer overcomes this by synthesizing novel 3D geometries and photorealistic textures entirely from scratch via diffusion models, enabling truly open-vocabulary scene diversity unconstrained by any asset library.

\subsection{Video Generation as Motion Priors}
The rapid progress in video generation~\cite{blattmann2023stable, wu2023tune} has sparked interest in using video models as world simulators or motion priors for robotics. UniSim~\cite{yang2023unisim} learns a universal interactive simulator from real-world action-annotated videos. Video language planning~\cite{cohen2024avdc} and UniPi~\cite{du2023learning} use video diffusion models to imagine future frames as a planning signal. GR-2~\cite{gu2024gr2} pre-trains on large-scale internet video and fine-tunes for manipulation, showing that video prediction priors enhance generalization. Genie~\cite{bruce2024genie} learns controllable world models from unlabeled play videos as interactive training grounds. Despite this progress, most approaches operate in 2D pixel space, making it difficult to extract precise, physically grounded 3D trajectories for robot control. V-Dreamer addresses this by employing a Sim-to-Gen alignment module based on CoTracker3~\cite{karaev2023cotracker} and VGGT~\cite{wang2025vggt} to lift 2D visual motion into physically executable end-effector trajectories, directly bridging visual synthesis and kinematic control.

\subsection{Sim-to-Real Transfer and Generalization}
Bridging the reality gap between simulation and the physical world remains a central challenge in robot learning. Domain Randomization (DR)~\cite{tobin2017domain, peng2018sim} randomizes physical and visual parameters to expose policies to diverse conditions, but requires manual tuning of randomization ranges and struggles to replicate real-world visual complexity. Real-to-Sim methods~\cite{zhang2023real2sim, torne2024rialto} reconstruct specific real scenes in simulation, but are labor-intensive and tied to particular environments. TRANSIC~\cite{jiang2023transic} refines simulation-trained policies via online human corrections, while DrEureka~\cite{ma2024dreureka} uses LLMs to automate randomization range design. V-Dreamer sidesteps these limitations by synthesizing a large variety of physically grounded scenes through generative diffusion models, naturally providing open-vocabulary visual diversity without manual specification or real-world interaction, and enabling zero-shot sim-to-real transfer to unseen objects.

%% file: section/4_method.tex










\section{Methodology}
\label{sec:method}

\textbf{V-Dreamer} is a fully automated framework designed to synthesize large-scale, physically grounded manipulation datasets for training generalizable robot policies. The core challenge in automated data generation is bridging the substantial representational gap between abstract human intentions and executable physical simulations. As illustrated in Figure~\ref{fig:pipeline}, V-Dreamer overcomes this challenge through a cohesive pipeline comprising three core modules: (1) \textit{Semantic-to-Physics Scene Synthesis}, which translates natural language into interactive 3D environments; (2) \textit{Video-Prior-Based Trajectory Generation}, which harnesses video foundation models to simulate expert manipulation behaviors; and (3) \textit{Sim-to-Gen Alignment}, which extracts actionable 3D trajectories from 2D visual predictions.

\subsection{Semantic-to-Physics Scene Synthesis}
\label{sec:scene_gen}

To transform an open-ended semantic instruction 
into a physically interactive rigid-body simulation, we propose a multi-stage procedural generation pipeline including: \ding{172} Semantic Parsing and Asset Manifest, \ding{173} Generative Visual Synthesis, \ding{174} Memory-Efficient 3D Reconstruction and \ding{175} Physically Grounded Layout. Specifically:

\ding{172} \textit{Semantic Parsing and Asset Manifest.} Given a user prompt, we first employ a Large Language Model (Qwen-Max\cite{qwen2025qwen25} in our experiment) acting as a high-level semantic planner. The LLM decomposes the abstract scene description into a structured \textit{Asset Manifest} in JSON format. This manifest explicitly specifies object categories and stylistic elements.
To ensure the downstream reconstruction quality, we inject specific negative constraints into the LLM prompt. For example, we prohibit highly reflective or transparent surfaces like mirrors and glass in a cyberpunk workspace, thereby enforcing style consistency and avoiding rendering artifacts across all generated assets. Note that this part only performs object-level manifest generation without background and layouts.

\ding{173} \textit{Generative Visual Synthesis.} For each manifest entry, we perform visual synthesis to obtain high-fidelity 2D image assets using the Flux diffusion model~\cite{flux2024}. To reduce environmental bias and avoid baked-in shadows which are critical for clean downstream 3D generation, we augment prompts with modifiers such as ``studio lighting'' and ``solid white background''. We then apply Segment Anything Model 3 (SAM3)~\cite{carion2025sam3} for precise background removal. For robustness, we adopt a hybrid segmentation scheme consisting of a semantic-first pathway that uses object names as prompts, and a vision-based fallback~\cite{gatis2023rembg} when semantic prompting produces suboptimal masks. With these clean, segmentation-ready 2D assets in place, the next step is to lift them into detailed 3D meshes for simulation. 

\ding{174} \textit{Memory-Efficient 3D Reconstruction.} Lifting high-resolution 2D assets to 3D meshes is often the primary computational bottleneck in an automated pipeline. To address this, we introduce a memory-efficient reconstruction pipeline. This module dynamically offloads encoder and decoder sub-networks between the CPU and GPU during inference. By strategically trading computation time for VRAM optimization, this mechanism enables the reconstruction of complex, high-poly geometries on standard consumer-grade hardware, outputting raw 3D meshes ready for physical instantiation.

\ding{175} \textit{Physically Grounded Layout.} After generating isolated 3D object meshes, the final step is to assemble them into a coherent and physically valid scene. We combine LLM spatial reasoning with rigorous physics-based validation. The LLM first infers approximate real-world metric dimensions for each object, after which we apply simple heuristics to correct distortions introduced during 3D lifting. Next, we generate the scene layout as a JSON specification using a foundation model, and then place each object at its corresponding 3D position accordingly. By this means, we construct a semantic scene graph that hierarchically categorizes objects into anchors and define the environment structure and accessories that serve as manipulable targets. Finally, we invoke a physics engine~\cite{Genesis} to enforce constraint-based layout via AABB collision checking and gravity alignment. Anchors are snapped to the ground plane ($Z=0$), while accessories are placed on supporting surfaces $Z=H_{\text{parent}}+\delta$ with procedural spatial jittering to systematically resolve bounding-box overlaps.

\subsection{Video-Prior-Based Trajectory Generation}
\label{sec:traj_gen}

Traditional motion planners often require task-specific engineering and may not directly benefit from the rich visual commonsense priors needed for open-vocabulary manipulation. In contrast, V-Dreamer exploits the rich, implicit motion priors encapsulated within video foundation models to synthesize diverse, context-aware manipulation trajectories.

Directly prompting a video model from a procedurally generated layout can lead to severe physical inconsistencies. Therefore, we first execute a brief dynamic settling phase within the Genesis~\cite{Genesis} simulator to resolve remaining micro-penetrations between objects. We capture this stabilized scene as the initial frame ($I_0$). We then perform style refinement on $I_0$ using Qwen-Image-Edit~\cite{wu2025qwenimagetechnicalreport} to obtain a \textit{Style-Refined Initial Frame} $I_0'$. This step is used in two ways: (i) The system automatically applies a set of pre-defined style perturbations to increase visual diversity, such as adding stripes or modifying material appearance. (ii) It provides a human-in-the-loop interface for targeted, user-specified stylization when customization is desired.

Building upon the stabilized frame, we feed $I_0'$ into the Wan2.2-I2V-Flash~\cite{wan2025} video generation model to synthesize a continuous temporal sequence of the specified manipulation task. To encourage physically plausible motion for robotic execution, we apply targeted negative prompts to suppress camera motion, object deformation, and background flicker. These linguistic constraints force the diffusion model to focus exclusively on the rigid-body kinematics of the manipulator and the target object, effectively suppressing the dream-like morphing artifacts that typically violate physical feasibility.

Since the synthesized video resides purely in the 2D pixel domain, the final step is to extract actionable 3D controls. We follow TraceGen~\cite{lee2026tracegen} and design a tracking-based grounding module. We first re-run SAM3 on the visually processed initial frame to obtain a high-quality mask of the target object. We then estimate per-frame depth using Visual Geometry Grounded Transformer (VGGT)~\cite{wang2025vggt}, which provides metric cues for 3D lifting. Next, we uniformly sample 2D points within the object mask and track them across frames using CoTracker3~\cite{karaev2023cotracker}, yielding dense but object-consistent 2D trajectories. These tracked points are subsequently fused into a coherent 3D motion trajectory using the TAPIP3D~\cite{tapip3d} module. Compared to TraceGen, which densely tracks pixels over the entire frame and follows moving regions, our mask-restricted tracking focuses on object-consistent motion and is empirically more precise. 

Finally, we apply lightweight post-processing to remove outlier tracks and suppress temporal jitter. Subsequently, we utilize Graspgen~\cite{murali2025graspgen} to generate a feasible robotic grasp pose for the target object, and then map the resulting 3D motion to the robot's end-effector trajectory via inverse kinematics (IK). This yields a physically executable trajectory $\tau = \{(p_t, q_t)\}_{t=0}^T$, where $p_t$ denotes the continuous end-effector pose and $q_t$ represents the discrete gripper state.

\begin{figure*}[t] 
    \centering 
    \includegraphics[width=0.95\linewidth]{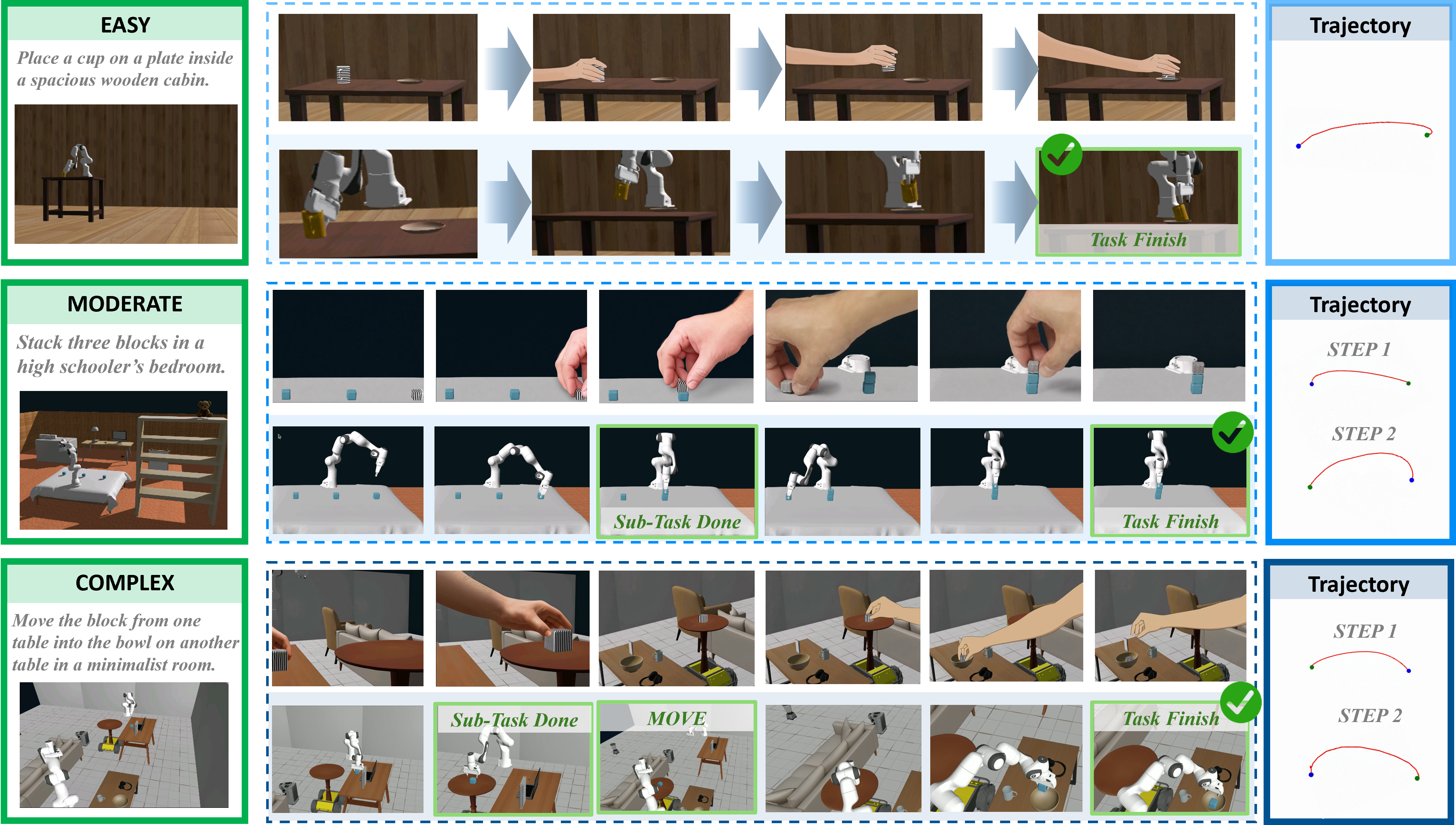} 
    \caption{Qualitative examples of V-Dreamer synthesized tasks of increasing difficulty. Left: Input Prompts and generated 3D scenes. Middle: Generated manipulation videos and the corresponding robot manipulation results. Right: Extracted end-effector trajectories for demonstrations.} 
    \label{fig:v_dreamer_demo} 
\end{figure*}

\subsection{Sim-to-Real Alignment}
\label{sec:sim2real}

By continuously running the V-Dreamer pipeline, we automatically generate a large-scale, high-fidelity synthetic dataset $\mathcal{D}=\{ {(\tau_i, O_i)}\}_{i=1}^{N} $, where each expert trajectory $\tau_i$ is paired with a diverse semantic and visual context $O_i$. For long instructions, the foundation model decomposes the instruction into several sub-tasks, and then iteratively performs video generation, trajectory generation and physical operation for each sub-task in simulation.

For physical deployment, we follow a strict sim-to-real transfer protocol. Specifically, we capture a snapshot of the real target scene and use it as an additional alignment input to instantiate a scene-matched, style-refined initial frame $I_0'$ in simulation. We also calibrate the simulator to match the real camera parameters and the robot workspace bounds. Conditioned on $I_0'$, all subsequent trajectory generation and policy training are performed entirely in simulation without any human demonstrations or target-domain fine-tuning. The policy trained solely on $\mathcal{D}$ is then deployed \textbf{zero-shot} on physical hardware for evaluation. 

For safety, we optionally reconstruct coarse bounding geometries of novel objects in simulation for background collision checking, while the core manipulation behavior and closed-loop control are driven entirely by the generalized representations learned from V-Dreamer’s synthesized data.

%% file: section/5_exp.tex
\section{Experiments}
\label{sec:experiments}

Experiments are designed to assess \textbf{V-Dreamer} as a robust data-generation engine by answering three core questions.

\textbf{Q1. Scalability and Data Quality.} Can V-Dreamer synthesize large-scale, high-quality demonstrations for downstream policy learning?

\textbf{Q2. Semantic and Geometric Diversity.} Can V-Dreamer generate sufficiently diverse object instances and spatial layouts to support zero-shot generalization to novel geometries?

\textbf{Q3. Physical Grounding and Executability.} Are V-Dreamer’s trajectories physically plausible and kinematically feasible, enabling zero-shot transfer on real hardware?

To the best of our knowledge, V-Dreamer is the first automated system-level framework that unifies open-vocabulary scene synthesis, video-prior-based trajectory generation, and real-world deployment. As prior works usually address only a single stage of this pipeline, there is no directly comparable end-to-end baseline. We therefore evaluate V-Dreamer from a system perspective, focusing on scene diversity, trajectory executability, policy learning utility, and sim-to-real transfer.

\begin{figure}[t!] 
    \centering 
    \includegraphics[width=0.9\linewidth]{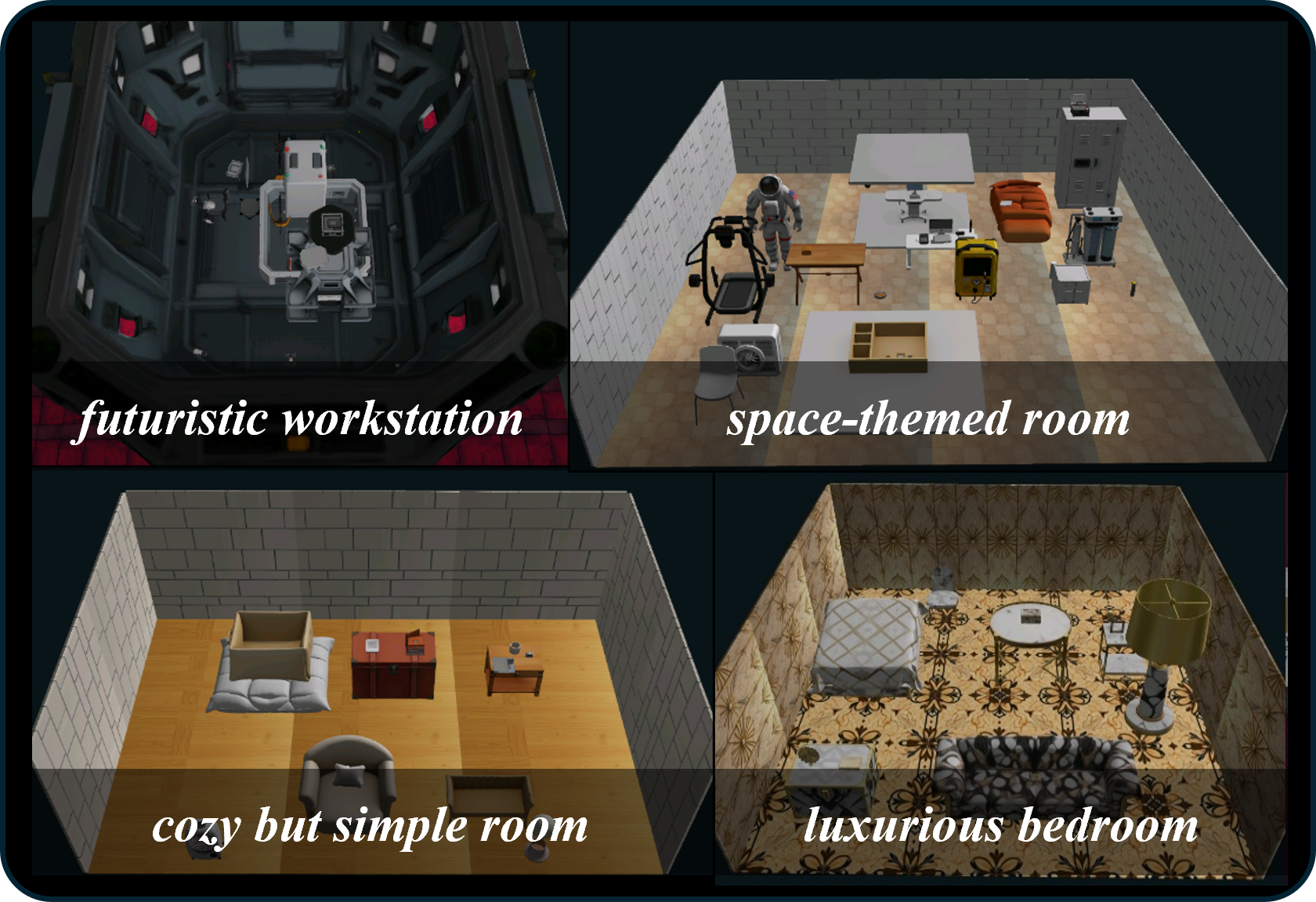} 
    \caption{Qualitative examples of V-Dreamer synthesized scenes with varying object instances, textures, and layouts. } 
    \label{fig:v_dreamer_scene} 
    \vspace{-2em}
\end{figure}

\begin{figure*}[t]
    \centering 
    \includegraphics[width=0.95\linewidth]{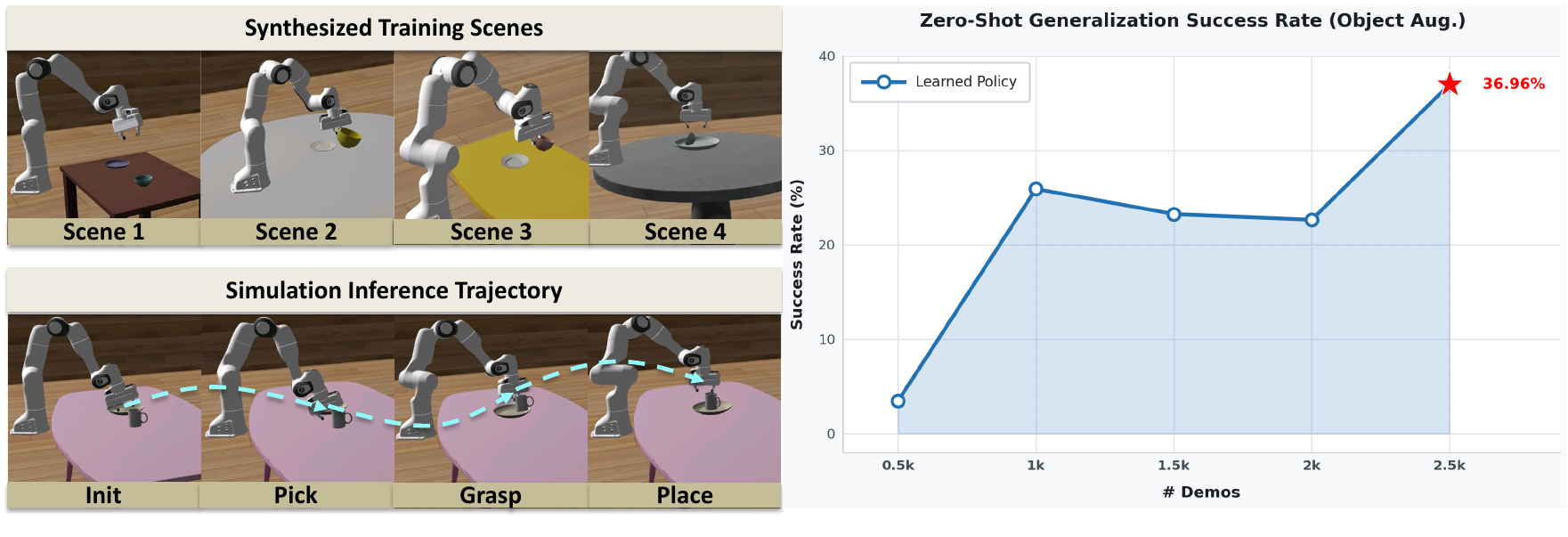} 
    \caption{
   \textbf{ Simulation Evaluation.} 
   \textit{(Left top)} Representative synthesized training scenes with object augmentation, illustrating diverse spatial layouts used to generate demonstrations. \textit{(Left bottom)} Zero-shot inference trajectory of the learned policy on an unseen mug, showing the manipulation sequence (Init $\to$ Pick $\to$ Grasp $\to$ Place). 
  \textit{ (Right)} Success rate of the learned policy on 10 unseen mug instances under different training dataset sizes (500 trials each). Performance improves with increasing data scale and peaks at 2.5k synthesized demonstrations.}
    \label{fig:sim_results} 
    \vspace{-1em}
\end{figure*}

\subsection{Experimental Setup}

\paragraph{Tasks and Dataset}
We systematically evaluate \textbf{V-Dreamer} on a fundamental yet challenging \textit{tabletop manipulation} task, where the robot grasps a target object from a planar surface and places it onto a receptacle. To support Q2, our pipeline procedurally synthesizes a diverse training corpus spanning (i) \textbf{Objects}: 40 target objects and 40 receptacle objects with substantial variation in geometry, texture, and metric scale; (ii) \textbf{Environments}: 20 distinct table settings with randomized surface materials, such as wood, marble, and glass. To evaluate zero-shot generalization, we additionally curate a held-out test set of 10 novel mugs with entirely unseen shapes, handles, and textures, which are strictly excluded from the generative pipeline.

\paragraph{Base Policy Model}
To evaluate whether data synthesized by V-Dreamer can effectively support robot policy learning, we adopt the classical imitation learning framework Action Chunking with Transformers (ACT)~\cite{zhao2023learning} as the downstream policy model. ACT is a widely used behavior cloning framework for robotic manipulation, providing a stable and reproducible training pipeline. Moreover, its sensitivity to demonstration quality and data distribution coverage makes it well suited for assessing the diversity and utility of synthesized data. Finally, since ACT does not involve exploration or reinforcement learning, performance differences can be more directly attributed to the training data itself.

\paragraph{Evaluation Metrics}
We use \textbf{Success Rate (SR)} as the evaluation metric, defined as the percentage of trials in which the robot successfully grasps and places the object within the episode horizon. Results are averaged over randomized trials across object instances and initial conditions, using the same metric for both simulation and real-world experiments.

\paragraph{Hardware Configuration}
In simulation, we use the 7-DoF Franka robotic arm. For real-world deployment, we evaluate the trained policies on the left arm of a Cobot-Magic platform, equipped with a 6-DoF Piper lightweight manipulator and a standard parallel-jaw gripper. For RGB perception, the system uses two Orbbec DaBai cameras: one mounted on the wrist and the other providing a top-down view. All synthetic training data are generated on a workstation equipped with 8$\times$RTX 4090 GPUs.


\subsection{Simulation Results}
\label{sec:sim_results}

\paragraph{Qualitative Visualization.}

To address \textbf{Q1}, we first visualize V-Dreamer’s synthesis capability. Figure~\ref{fig:v_dreamer_demo} further shows that V-Dreamer can generate coherent demonstrations across increasing levels of interaction difficulty, from \textit{simple} motions to \textit{hard} and \textit{complex} cooperative sequential mobile manipulation task by three devices by presenting both the synthesized videos and the recovered end-effector trajectories. These examples indicate that V-Dreamer produces structured, executable scene-action pairs rather than merely randomizing assets. As a supplement, we show more generated scenes with substantial variation in texture, metric scale, and spatial layout in Figure~\ref{fig:v_dreamer_scene}.

\paragraph{Data Scaling and Zero-Shot Generalization}
We further answer \textbf{Q2} and evaluate \textbf{V-Dreamer} as a data-generation engine along two axes, including (i) whether adding more synthesized demonstrations improves downstream learning, and (ii) whether the learned policy transfers to geometrically unseen objects. Concretely, we train the same diffusion-based imitation policy on subsets of the synthesized dataset ranging from 500 to 2,500 trajectories, and evaluate each variant on a held-out set of 10 novel mugs, conducting 50 trials per object with randomized initial configurations.

The left side of Figure~\ref{fig:sim_results} presents qualitative results of execution trajectories in simulation, where the policy is trained with synthesized data and evaluated on unseen mug instances. The quantitative results, summarized in the right side of Figure~\ref{fig:sim_results}, reveal a non-linear correlation between generative data scale and policy success rate. The policy exhibits catastrophic failure when trained on only 500 trajectories (3.46\%), indicating an insufficient coverage of the visual-spatial state distribution. Performance leaps significantly at 1,000 trajectories (25.90\%) and peaks optimally at \textbf{2,500 trajectories} with a success rate of \textbf{36.96\%}. 
Notably, this peak is achieved using purely synthesized demonstrations on entirely unseen mug geometries, highlighting V-Dreamer’s ability to generate diverse, actionable training data. These results empirically confirm that V-Dreamer provides high-quality data for robust manipulation primitives, and given that current imitation learning methods rarely exceed 30\% success on unseen tasks~\cite{chen2025robohiman}, this further demonstrates its effectiveness for imitation learning.


\subsection{Real-World Experiments}

\begin{figure}
    \centering
    \includegraphics[width=0.95\linewidth]{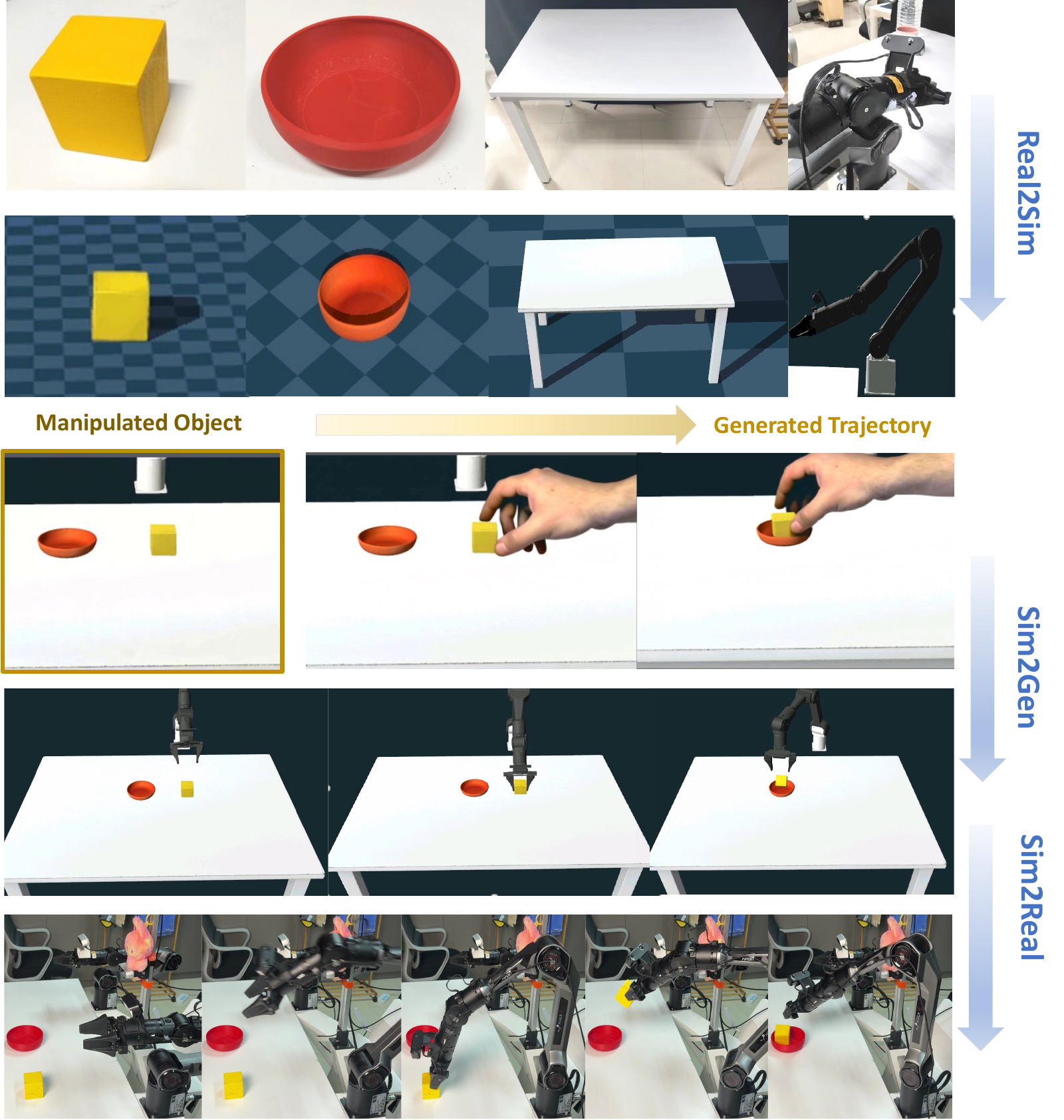}
    \caption{\textbf{V-Dreamer enables a Real2Sim–Sim2Gen–Sim2Real pipeline for real-world robot manipulation.} Real-world observations are first converted into simulation environments (Real2Sim). V-Dreamer then synthesizes executable manipulation trajectories in simulation (Sim2Gen) for policy training. Finally, the learned policy is deployed on a physical robot arm (Sim2Real), successfully executing a pick-and-place task that transfers a novel yellow cube to the red target without any real-world fine-tuning.}
    \label{fig:real_vis}
    \vspace{-1em}
\end{figure}





To evaluate the extreme data efficiency and practical utility of \textbf{V-Dreamer}, we deploy the learned policy on a physical 6-DoF Piper robotic arm (left arm of an ALOHA setup). Unlike conventional methods requiring large-scale real demonstrations, we follow a strict sim-to-real protocol. The policy is trained on a single \textbf{one-shot} V-Dreamer trajectory in simulation, generated for a simple scene with a block, a bowl, and a table, and then transferred to the real world in a \textbf{zero-shot} manner.

We adopt a minimalist zero-shot sim-to-real setup for physical deployment. For low-cost domain alignment, we use the Piper URDF from Robot-Twin, refine joint limits, render the simulated arm links (except the base) in black, and apply black tape to the corresponding real links within the camera view to reduce robot-body appearance mismatch. The policy runs closed-loop using observations from an Orbbec DaBai RGB-D camera, while we intentionally feed only the raw RGB image as $O_t$ and discard depth and extra calibration to test robustness. The full pipeline is illustrated in Figure~\ref{fig:real_vis}. 

\begin{table}[htbp]
\begin{threeparttable}
\centering
\caption{One-Shot Sim-to-Real Robustness Evaluation}
\label{tab:real_world_robustness}
\begin{tabular}{@{}ccc@{}}
\toprule
\centering
\textbf{Perturbation Type} & \textbf{Real-World Setup} & \textbf{SR} \\ \midrule
Visual Distractors & Unmodeled background clutter in FOV & 50\% \\
Object Generalization & Apple, tape, mango, pear, bottle & 20\% \\
Spatial Perturbation & Shifted initial object and goal positions & 15\% \\
Sensor Occlusion & Blocked Orbbec DaBai camera view & 0\% \\ \bottomrule
\end{tabular}

\begin{tablenotes}
\item All tests used a policy trained on a single synthetic trajectory. The 0\% success rate under occlusion confirms reliance on closed-loop visual feedback rather than open-loop playback.
\end{tablenotes}

\end{threeparttable}
    \vspace{-1em}
\end{table}

We evaluate four perturbation conditions on a one-shot policy trained using a single V-Dreamer-synthesized trajectory, focusing on Success Rate (SR). As shown in Table~\ref{tab:real_world_robustness}, the conditions include: visual distractors on the table (50\% success), out-of-distribution target objects such as apple, tape roll, mango, pear, and bottle (20\% success), spatial perturbations of both target and receptacle poses (15\% success), and camera occlusion during execution (0\% success). The occlusion result confirms true closed-loop visual servoing rather than open-loop trajectory replay, while the other conditions highlight non-trivial robustness and limited generalization under extreme one-shot supervision.

These real-world results primarily validate \textbf{V-Dreamer} as a data synthesis engine, rather than an algorithmic tweak aimed at maximizing task accuracy. Under a strict zero-shot sim-to-real protocol, a policy trained on a \emph{single} V-Dreamer-synthesized demonstration can execute meaningful behaviors on physical hardware. This indicates that the generated trajectories are physically grounded, kinematically feasible, and directly usable for control, providing direct evidence for \textbf{Q3}. V-Dreamer distills generative video priors into executable demonstrations, bridging the reality gap even in an extreme low-data regime.


%% file: section/6_conclusion.tex
\section{Conclusion}
\label{sec:conclusion}
In this work, we presented \textbf{V-Dreamer}, a fully automated framework for synthesizing large-scale, physically grounded simulation environments and executable manipulation trajectories for generalist robot learning. By integrating Large Language Models with advanced 2D/3D and video generative models, V-Dreamer translates open-vocabulary instructions into diverse interactive scenes and coherent manipulation behaviors. Unlike prior generative pipelines that primarily focus on scene synthesis, V-Dreamer tackles the full generation loop, producing training-ready data that is directly usable for policy learning. 

Extensive experiments show that imitation policies trained exclusively on V-Dreamer-generated data achieve robust performance and zero-shot generalization to geometrically unseen objects in simulation. Furthermore, with photo-conditioned scene alignment, these policies transfer zero-shot to real hardware. These results suggest that the trajectories generated by V-Dreamer are not merely visually plausible, but physically executable and directly useful for robot control, highlighting the promise of full-cycle generative pipelines in alleviating the data bottleneck in robotic manipulation.

\textbf{Limitations and Future Work.}
An important next step is to develop automated, physics-aware trajectory filtering to reduce the impact of low-quality generations. In addition, the current framework is limited to rigid-body tabletop manipulation. Extending it to articulated and deformable objects is a key direction for future work. Finally, incorporating tighter physical feedback during generation and grounding may further improve physical realism and sim-to-real robustness.